\documentclass[11pt]{article}

\usepackage[final]{acl}

\usepackage{times}
\usepackage{latexsym}

\usepackage[T1]{fontenc}
\usepackage[utf8]{inputenc}

\usepackage{microtype}

\usepackage{inconsolata}

\usepackage{graphicx}
\usepackage{booktabs}
\usepackage{multirow}
\title{$Mem^p$: Exploring Agent Procedural Memory}

\author{
  Runnan Fang\textsuperscript{$\spadesuit\heartsuit*$}, 
  Yuan Liang\textsuperscript{$\spadesuit$}\thanks{$\quad$ Equal Core Contributors.}, 
  Xiaobin Wang\textsuperscript{$\heartsuit$},
  \textbf{Jialong Wu}\textsuperscript{$\heartsuit$},\\
  \textbf{Shuofei Qiao}\textsuperscript{$\spadesuit\heartsuit$}, 
  \textbf{Pengjun Xie}\textsuperscript{$\heartsuit$},  
  \textbf{Fei Huang}\textsuperscript{$\heartsuit$}, 
  \textbf{Huajun Chen}\textsuperscript{$\spadesuit$}, 
  \textbf{Ningyu Zhang}\textsuperscript{$\spadesuit\clubsuit$}\thanks{$\quad$ Corresponding Author.}  \\
   \textsuperscript{$\spadesuit$}Zhejiang University 
   ~$^\heartsuit$Alibaba Group  \\
   \textsuperscript{$\clubsuit$}State Key Lab. for Novel Software Technology,\\
   Nanjing University, P.R. China \\
  \texttt{\{rolnan,zhangningyu\}@zju.edu.cn} \\
    \vspace{.5em}%\github
    \href{https://github.com/zjunlp/MemP}{Code: https://github.com/zjunlp/MemP}
}
\begin{document}
\maketitle
\begin{abstract}
Large Language Models (LLMs) based agents excel at diverse tasks, yet they suffer from brittle procedural memory that is manually engineered or entangled in static parameters. In this work, we investigate strategies to endow agents with a \textit{learnable}, \textit{updatable}, and \textit{lifelong} procedural memory. We propose $Mem^p$ that distills past agent trajectories into both fine-grained, step-by-step instructions and higher-level, script-like abstractions, and explore the impact of different strategies for \textit{Build}, \textit{Retrieval}, and \textit{Update} of procedural memory. Coupled with a dynamic regimen that continuously updates, corrects, and deprecates its contents, this repository evolves in lockstep with new experience. Empirical evaluation on TravelPlanner and ALFWorld shows that as the memory repository is refined, agents achieve steadily higher success rates and greater efficiency on analogous tasks. Moreover, procedural memory built from a stronger model retains its value: migrating the procedural memory to a weaker model can also yield substantial performance gains.
\end{abstract}

\section{Introduction}
As large language models (LLMs) grow ever more powerful, LLM-based agents augmented by their own reasoning and external tools are taking on increasingly sophisticated works \cite{zhao2023survey,wang2024survey,xi2025rise,reasoning-survey}.
No longer mere assistants, these agents now trawl the web for elusive insights and weave them into comprehensive, publication ready reports, like Deep Research~\cite{dr,grok} and WebDancer~\cite{intro/webdancer}. 
Moreover, they can handle complex data analyses~\cite{intro/dasurvey, intro/damethod, intro/automind}, navigate multi-step GUI workflows~\cite{intro/gui,intro/uitars}, and sustain long-horizon, tool-rich interactions~\cite{intro/tau-bench,intro/tau2-bench,intro/acebench, fang2025synworld,gur2023real} with precision.
Yet executing such intricate, long-horizon tasks demands dozens of steps and protracted runtimes. 
Along the way, unpredictable external events—network glitches, UI changes, shifting data schemas—can derail the entire process. 
Restarting from scratch every time is a punishing ordeal for present-day agents. 
Beneath their surface diversity, many complex tasks share deep structural commonalities and a similar environment. 
Instead of starting fresh each time, an agent should extract its experience from past successes. 
By turning earlier trajectories into reusable templates like patterns of reasoning, tool sequences, and recovery tactics, it can progress step by step, learning from every failure and success, until even the most convoluted missions become routine.

\begin{figure}[!t] % !t, !htbp
    \centering
    \scalebox{1}{
    \includegraphics[width=1\linewidth]{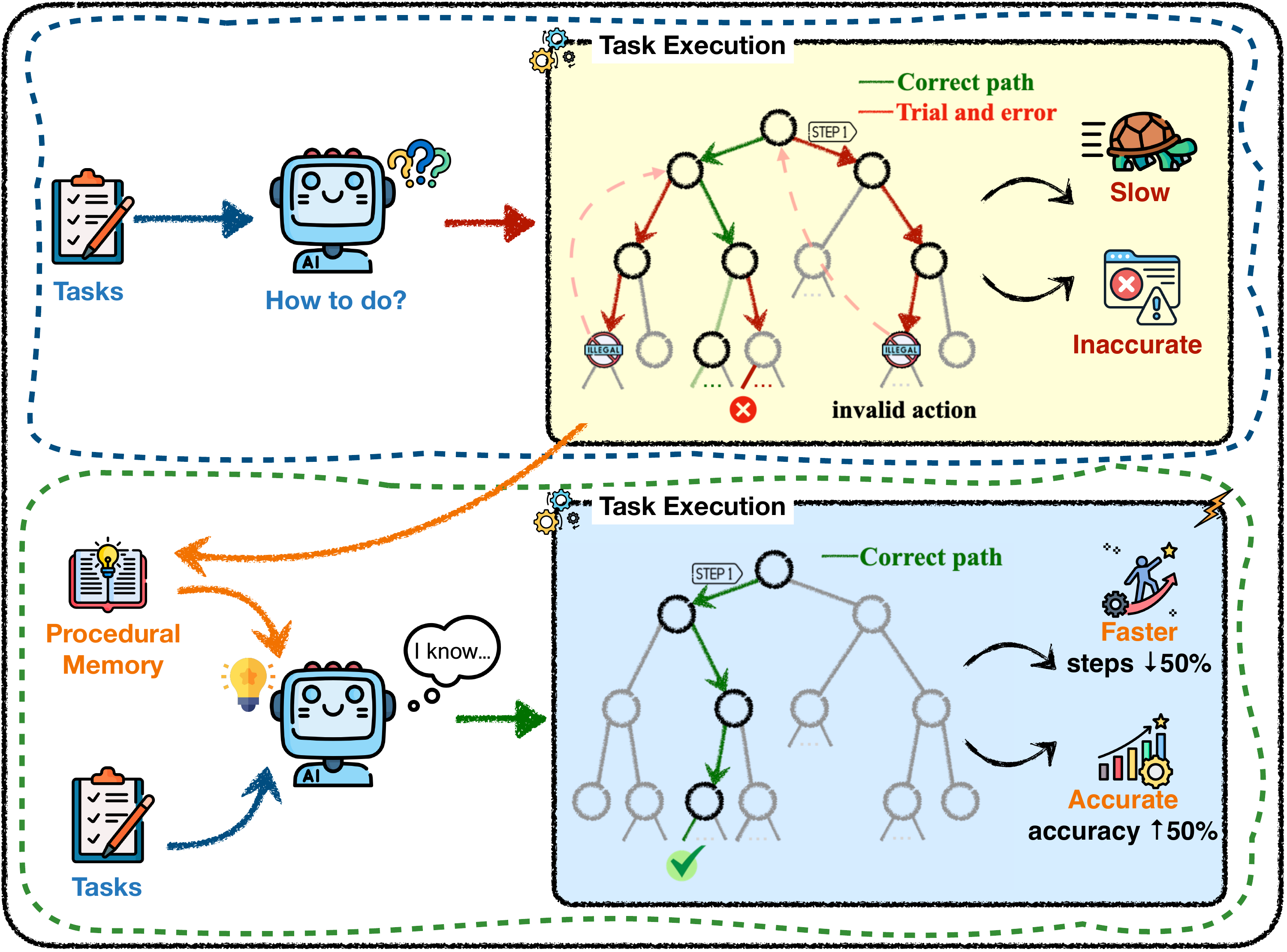} 
    }
    \caption{With \textbf{procedural memory}, agents can improve both the success rate (accuracy ↑) and execution efficiency (steps ↓) when solving similar tasks.}
    \label{fig:intro}
    % \vspace{-1mm}
\end{figure} 
The capacity to distill, chronicle, and re-apply lessons from one’s own experiential trajectory is the bedrock of human learning and the pivotal gateway through which an agent ascends toward self-directed refinement~\cite{liu2025advances,sumers2023cognitive,li2023camel}. 
Procedural memory~\cite{intro/proceduralmem,intro/1980preserved} silently compiles habitual skills into executable subroutines, enabling unconscious, fluent action. 
While contemporary agents built on LLMs can compose short action plans or call external tools, their procedural knowledge is either hand-crafted, stored as brittle prompt templates, or implicitly entangled in model parameters that are expensive to update. 
Existing memory-augmented frameworks such as LangGraph~\cite{intro/langgraph}, AutoGPT~\cite{intro/autogpt}, or agent cognitive architectures like Memory Bank~\cite{intro/memorybank,cognitive} and Soar~\cite{intro/soar} provide coarse abstractions (buffers, rule chunks, production systems) but leave the optimization of procedural memory life-cycle operations about how skills are built, indexed, patched, and eventually pruned, largely unexamined. 
Consequently, there is no reliable way to measure how efficiently an agent grows its procedural skills or to ensure new experiences improve rather than harm performance.

To close this gap, we present $Mem^p$, a task-agnostic framework that treats procedural memory as a first-class optimization object.
The core exploration of $Mem^p$ lies in how different strategies for memory construction, retrieval, and updating affect overall performance. During the construction phase, we follow the majority of traditional memory architectures and agent-based memory designs by leveraging either the full historical trajectory or explicit guidelines to guide the process. In the retrieval phase, we experiment with various key-building strategies—such as query-vector matching and keyword-vector matching—to investigate how procedural memory can be constructed more precisely.
Unlike prior memory mechanisms or learning from experience, $Mem^p$ introduces diverse procedural-memory update strategies:
In the realm of agents, memory updating is crucial for agents to adapt to dynamic environments. By incorporating diverse strategies like ordinary addition, validation filtering, reflection, and dynamic discarding, agents can efficiently manage their knowledge base. This ensures they stay updated with new information, discard outdated data, and optimize memory resources. Such strategies enhance learning efficiency, improve decision-making quality, and boost adaptability, allowing agents to perform optimally in various tasks and scenarios.

% 实验概述以及最终结果=
We instantiate $Mem^p$ on top of strong LLMs (GPT-4o and Claude, Qwen) and evaluate on two diverse domains: long-horizon housework ALFWorld~\cite{alfworld} and long-term information seeking task TravelPlanner~\cite{travelplanner}.
On two benchmark datasets that rigorously evaluate agent capabilities, we demonstrate that constructing and retrieving procedural memory during training empowers an agent to distill and reuse its prior experience. When this memory is exploited at test time, the agent’s task accuracy rises, and compared with tackling each instance in isolation, it eliminates most fruitless exploration on unfamiliar tasks, yielding substantial reductions in both step count and token consumption.
Further, by equipping the agent with a set of memory-update mechanisms, we allow it to build and refine its procedural memory while acting in the test environment. This endows the agent with a continual, almost linear mastery of the task. %Extensive ablations reveal that procedural memory also scales gracefully and transfers effectively to new, related tasks
Extensive ablation studies reveal that procedural memory not only scales gracefully with increasing task complexity but also transfers effectively to new, related tasks, demonstrating its adaptability across scenarios.

\section{Related Works}
\paragraph{Memory in Language Agents.}
Memory is a foundational component in language agents, enabling them to retain and utilize past information across multiple timescales, including short-term, episodic, and long-term memory, to enhance their performance and adaptability~\cite{zhou2023agents,zhou2024agents2,zhang2024survey,liu2025advances,memos}.
These systems aim to mimic aspects of human memory to improve coherence, personalization, and learning capabilities ~\cite{mem0,wu-etal-2025-webwalker,minerva}.
% Memory allows language agents to maintain context in conversations, learn from past experiences, adapt to changing environments, and provide personalized interactions. 
% Without memory, agent systems would lack the ability to build on previous knowledge and would be limited to stateless, reactive behaviors.
Current approaches include end-to-end memory systems~\cite{yu2025memagent,zhou2025mem1}, external memory systems~\cite{mem0,zhong2024memorybank}, and hierarchical memory structures~\cite{hu2024hiagent,xu2025mem}. 
These methods involve encoding and storing information in various formats, using retrieval mechanisms like vector embeddings and semantic search, and implementing memory updating and forgetting strategies to maintain relevance and efficiency.
% However, most works do not focus on memory in multi-turn agent interactions. 
% How to enable agents to learn memory across trajectories remains a challenge.
Despite its importance, memory in multi-turn agent interactions remains underexplored, and enabling agents to effectively learn and utilize memory across trajectories poses a significant challenge.
Procedural memory is a type of long-term memory that involves the retention of procedures and skills, such as  typing or riding a bike, which are performed automatically without conscious thought. 
The agent utilizes procedural memory to internalize and automate repetitive tasks, decision-making processes, and interaction patterns, leading to more efficient and context-aware responses over time. 
Although there have been several works, such as Voyager~\cite{wang2023voyager}, AWM~\cite{wang2024agent}, and AutoManual~\cite{chen2024automanual}, that utilize procedural memory to enhance agents' capabilities on similar tasks, there still lacks a systematic analysis on how to construct, retrieve, and update such procedural memory like ~\cite{memoryupdating}. 
Therefore, our work mainly focuses on exploring how to build an effective procedural memory system for agents performing cross-trajectory tasks.

%\vspace{-8pt}
\paragraph{Learning from Experience.}
LLM-based Agent learning from experience involves intelligence continuously improving their decision-making capabilities through interaction with environments and utilization of past experiences~\cite{Meta-Agent-Workflow,tang2025agentkb,zhou2025mem1,worfbench,learn-by-interact,wang2024agent}.
This approach is crucial for developing adaptive and intelligent agents capable of handling dynamic real-world scenarios, as it allows them to optimize behaviors, reduce manual programming needs, and enhance performance across various tasks~\cite{zhengsynapse, rw/liu2025ml, rw/mobileagent}. 
Agents typically employ mechanisms such as reinforcement learning~\cite{rw/lu2025arpo,rw/toolstar}, experience replay~\cite{rw/feng2025get,rw/liu2025contextual}, imitation learning~\cite{rw/agentimitation,rw/agentinit}, memory management~\cite{rw/hou2024my,rw/hu2024hiagent}, and multi-agent learning to achieve this. 
However, current methods face limitations including low sample efficiency, poor generalization across tasks, catastrophic forgetting when learning new information, and there are few features for memory update. 
The key distinction of our work lies in systematically investigating optimal strategies for construction, retrieval, and update modules of an agent’s procedural knowledge. 
During the update phase, we enhance the agent’s capabilities by maintaining an editable repository of procedural knowledge.
Additionally, collecting high-quality training data can be challenging and may introduce biases. 
Addressing these limitations is essential for advancing the capabilities of LLM-based agents and ensuring their effective application in real-world contexts.

\section{Preliminary}
\begin{figure*}[!t] % !t, !htbp
    \centering
    \scalebox{1}{
    \includegraphics[width=1\linewidth]{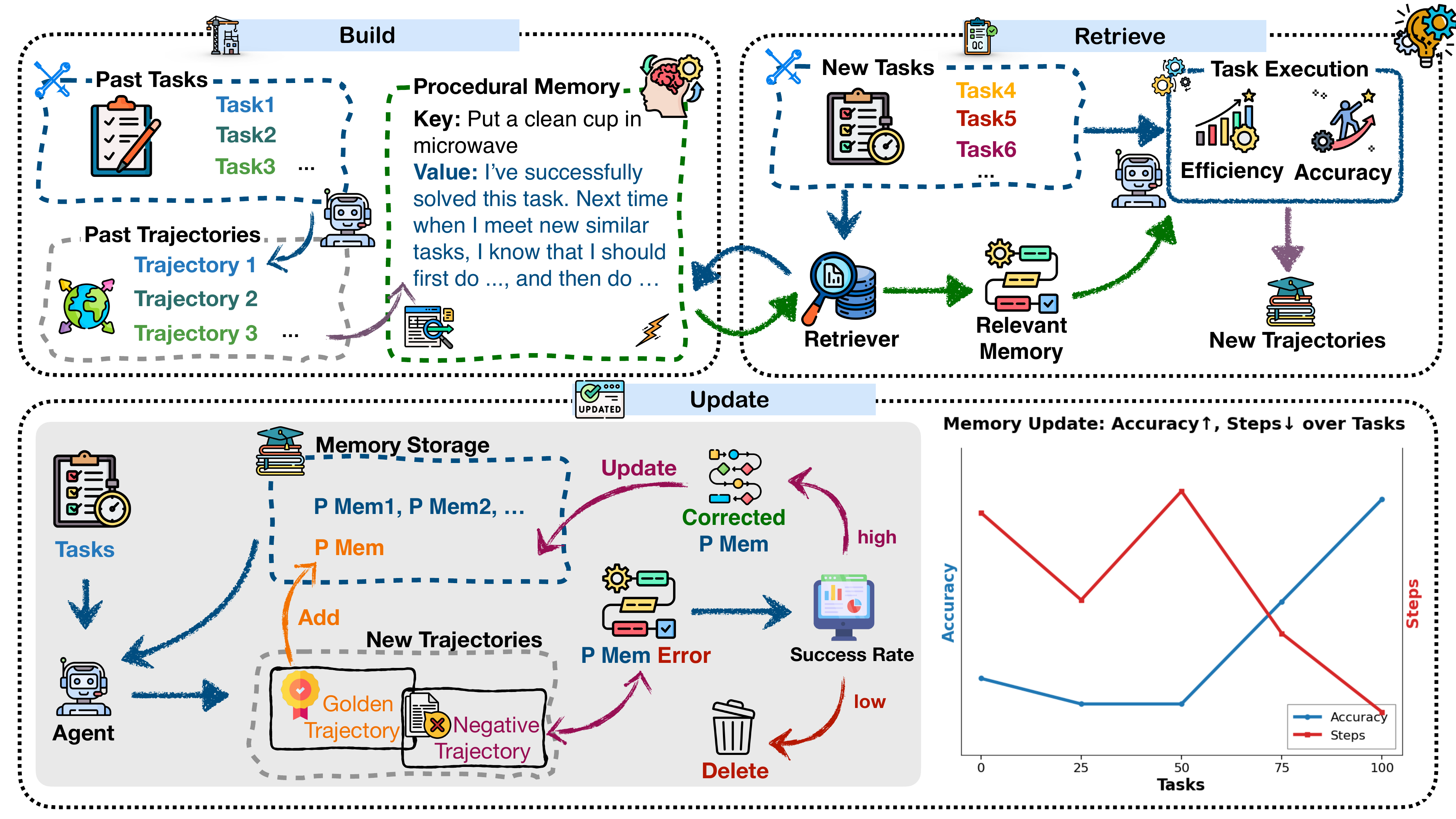} 
    }
    \caption{The procedural memory framework consists of \textbf{Build}, \textbf{Retrieve}, and \textbf{Update}, which respectively involve encoding stored procedural memory, forming new procedural memories, and modifying existing ones in light of new experiences.}
    \label{fig:framework}
    % \vspace{-1mm}
\end{figure*} 

When an agent influences its external environment by invoking external tools or executing prescribed actions, and iteratively refines its behavior over multiple rounds to accomplish a complex multi-step objective, this paradigm can be modeled as a Markov Decision Process (MDP).~\cite{puterman1990markov}
Under this view, at each discrete time step $t$, the agent, situated in state $s_t \in S$, chooses an action $a_t \in A$, according to its policy $\pi(a_t|s_t)$, where $A$ is the action space of the task.  
The environment then transitions to a new state $s_{t+1} \in S$ and emits an observation $O_t$.  
Consequently, the entire interaction trajectory may be compactly expressed as:
\begin{equation}
\tau = (s_0, a_0, o_1, s_1, a_1, o_2, \ldots, s_T),
\end{equation}
where $\tau$ is the complete exploration trajectory of this task.
Moreover, a reward function $R$ will evaluate the task's completion $r$ within this environment $env$ by assigning a score based on the final state $s_T$ or the entire trajectory $\tau$. 
\begin{equation}
r = R(env, s_T, \tau)\in[0,1]
\end{equation}

% \begin{equation}
% \tau = (s_0, a_0, o_1, s_1, a_1, o_2, \ldots, s_T)
% \end{equation}

Although approaches resembling Markov Decision Processes inevitably contain erroneous actions and exploratory attempts, the contextual information they generate becomes valuable for decision-making as the model’s reasoning and reflective capabilities improve. Nevertheless, this benefit comes at a high test-time cost—both in time and in token consumption. When facing an entirely new and complex environment, many actions (or tokens) are spent simply understanding the environment and the task itself. This leads to redundancy when similar tasks are executed within the same environment: the agent has already acquired partial procedural knowledge about the environment or task during earlier episodes, yet fails to transfer that knowledge effectively to subsequent tasks.

By shifting from parallel task completion to sequential task completion, the agent can learn and distill experience from earlier tasks, thereby reducing repetitive exploration. Inspired by human procedural memory, we propose to equip the agent with a procedural memory module. This module transforms the conventional policy $\pi(a_t | s_t)$ into $\pi_{m^p}(a_t | s_t)$, where $m^p$ is the agent’s learned procedural memory.

\subsection{Agent Procedural Memory}
Procedural memory is the type of long-term memory responsible for knowing how to perform tasks and skills, such as typing or riding a bike. 
By mastering this type of procedural memory, humans avoid the need to relearn the process each time.  
For an agent, that is, for a task trajectory $\tau$ and its reward $r$, a memory $m^p$ is constructed by a builder $B$, thereby achieving the acquisition of memory, namely
\begin{equation}
    Mem = \sum_{t=1}^{T} m^{p_t}, where\ m^{p_t} = B(\tau_t,r_t)
\end{equation}
where $Mem$ is the procedural memory library acquired by the agent over the $T$ tasks.
After constructing the procedural memory library, when facing a new task $t_{new}$, we need a good procedural memory retriever to recall a memory that fits $t_{new}$. Generally speaking, we would choose the task $t 
\in T$ that is most similar to $t_{new}$, because similar experiences are more helpful for the agent to complete the new task.
\begin{equation}
m_{retrieved} = \arg\max_{m^{p_i} \in Mem} S(t_{new}, t_i)
\end{equation}
As we use cosine similarity for the vector embedding model $\phi$ of the task in the experiment, the retrieval process becomes:
\begin{equation}
    m_{retrieved} = \arg\max_{m^{p_i} \in Mem} \frac{\phi(t_{new}) \cdot \phi(t_i)}{\|\phi(t_{new})\| \|\phi(t_i)\|}.
\end{equation}
Moreover, as the number of completed tasks continuously increases, simply augmenting the agent's procedural memory is inconsistent with common sense. A well-designed procedural memory system should have a reasonable update mechanism—that is, it should dynamically perform addition, deletion, modification, and retrieval based on the task execution context.

Let $M(t)$ denote the agent's procedural memory at time $t$, and $\tau_t$ represent the set of tasks completed up to time $t$. Then, the update mechanism can be modeled as a function $U$ that takes the current procedural memory and task execution feedback to produce the updated memory:
\begin{equation}
    M(t+1) = U\left( M(t), E(t), \tau_t \right),
\end{equation}
where $E(t)$ encapsulates the execution feedback (e.g., success, failure, performance metrics). A more sophisticated implementation of $U$ could be represented as:

\begin{equation}
U = Add \left( M_{new} \right) \ominus Del\left( M_{obs} \right) \oplus Update(M_{est})
\end{equation}
where $M_{new}$ represents new procedural memory to be added; $M_{obs}$  indicates procedural memory to be removed, $M_{est}$ are tasks to be updated based on execution feedback $E(t)$. This comprehensive formula captures the essential add, delete, and modify operations within the update mechanism.

\section{Experiment}

\begin{table*}[h]
  \centering
  \scalebox{0.86}{
  \begin{tabular}{c|l|ccc|ccc}
  \toprule
  \multirow{2}{*}{\textbf{Model}} & \multirow{2}{*}{\textbf{Granularity}} & \multicolumn{3}{c|}{\textbf{TravelPlanner}} & \multicolumn{3}{c}{\textbf{ALFWorld}} \\
  & &  {\#\textit{CS} ↑} & {\#\textit{HC} ↑} & {Steps ↓} & {Dev ↑} & {Test ↑} & {Steps ↓} \\
  \midrule
  \multirow{4}{*}{\textbf{GPT-4o}}
  & No Memory & 71.93 & \textbf{12.88} & 17.84 & 39.28 &  42.14 & 23.76 \\
  & Script & 72.08 & 5.50 & 15.79 & 66.67 & 56.43 & 18.52 \\
  & Trajectory & \underline{76.02} & 8.25 & \underline{14.64} & 67.17 & 74.29 & \underline{16.49} \\
  & Proceduralization & \textbf{79.94} & \underline{9.76} & \textbf{14.62} & \textbf{87.14} & \textbf{77.86} & \textbf{15.01} \\
  \midrule
  \multirow{4}{*}{\textbf{Claude-3.5-sonnet}}
  & No Memory & 63.49 & \textbf{33.06} & 18.84 & 39.20 & 34.97 & 24.12 \\
  & Script & 62.08 & 29.61 & 19.21 & 56.13& 53.59& 19.38 \\
  & Trajectory & \underline{65.76} & 29.61 & \underline{17.72} & \underline{69.28} & \underline{71.78} & \underline{15.97} \\
  & Proceduralization & \textbf{65.46} & \underline{30.14} & \textbf{15.29} & \textbf{82.50} & \textbf{74.72} & \textbf{15.79} \\
  \midrule
  \multirow{4}{*}{\textbf{Qwen2.5-72b}}
  & No Memory & 56.57 & 7.34 & 18.32 & 44.91 &41.25 & 21.38\\
  & Script & 58.59 & 7.34 & 18.53 & \underline{66.24} & 61.88&17.13 \\
  & Trajectory & \underline{63.41} & \underline{12.66} & \underline{18.12} &64.49& \underline{69.57} & \underline{16.40} \\
  & Proceduralization & \textbf{63.82} & \textbf{14.19} & \textbf{17.94} & \textbf{85.71} & \textbf{77.19} & \textbf{15.32} \\
  \bottomrule
  \end{tabular}
  }
  \caption{Results on \textbf{Build Policy}. 
  \#\textit{CS}, \#\textit{HC} denote Commonsense and Hard Constraint, respectively. 
  ↑ indicates the higher values are better, and ↓ denotes the lower values are better.
  The best results among all methods with similar settings are \textbf{bolded}, and the second-best results are \underline{underlined}.}
  \label{table: Storage}
\end{table*}

In this section, we will introduce the Procedural Memory framework in detail (Figure~\ref{fig:framework}), covering the storage, retrieval, and update modules of memory, as well as analyzing which strategies perform better within each module.
\subsection{Experimental Settings}

\paragraph{Datasets.}
For our experiments, we adapt TravelPlanner~\citep{travelplanner} and ALFWorld~\citep{alfworld} benchmarks. 
TravelPlanner is a benchmark designed to evaluate agents’ ability to use tools and perform complex planning under intricate constraints. 
In contrast, ALFWorld comprises household tasks.
In each interaction round, the agent outputs an action, and the environment responds with textual feedback describing the resulting state. 
This process repeats for multiple turns until the task is completed or the maximum number of rounds is reached.
ALFWorld includes test split to evaluate the agent’s generalization ability. Detailed evaluation methods are shown in Appendix~\ref{app: evaluation}.
\paragraph{Backbones.}
% 在实验中我们选择了两个强baseline 模型, gpt-4o 以及claude, 以及一个开源的强模型 qwen2.5-72b-instruct. 在实验中 我们重要
In our experiments, we benchmarked our procedural memory on three base models. 
Specifically, we adopt the two proprietary frontier models that have consistently dominated public leaderboards: OpenAI’s GPT-4o~\cite{gpt4} and Anthropic’s Claude~\cite{claude}, and complement them with the open-sourced Qwen2.5-72B-Instruct~\cite{yang2024qwen2}. 
The first two provide state-of-the-art closed-source performance, while the third allows us to verify that our findings generalize beyond proprietary systems and remain valid in the open-source regime.

\subsection{Memory Storage \& Retrieval}

Procedural knowledge is typically stored in two main formats: (1) trajectories are kept verbatim, round by round, in memory, or (2) high-level abstractions are extracted from these trajectories and then stored. 
Once a similar procedural memory is retrieved, it is appended to the task as part of the context, serving as prior knowledge to assist the model in completing the task. 

Inspired by this, we designed the following experimental conditions:
\begin{itemize}
  \item \textbf{No Memory:} The model tackles the assigned task in a ReAct fashion without any external memory.  
  \item \textbf{Trajectory:} We first filter the gold trajectories from the training set and store them. At inference time, the system retrieves the top-k trajectories whose query vectors are most similar to the current task’s vector, supplying them as procedural memories before execution.  
  \item \textbf{Script:} The model analyzes and summarizes the gold trajectories from the training set, distilling them into abstract procedural knowledge that is provided as a prompt before each task.  
  \item \textbf{Proceduralization:} This condition combines the full retrieved trajectories with the high-level script generated by the model, integrating both concrete examples and abstract guidance as the procedural memory.
\end{itemize}

As shown in Table~\ref{table: Storage}, all memory construction methods outperform the no-memory baseline, achieving higher scores on both datasets while also reducing the number of steps required.
This indicates that procedural memory built during training is beneficial for directly applying tasks during testing. Furthermore, we observe that the approach of abstracting trajectories into scripts during training yields relatively better performance on the ALFWorld test set compared to the dev set. 
Conversely, trajectories that utilize complete execution traces as procedural memory achieve higher scores on the dev set, suggesting that scripts are more capable of generalizing to different test tasks, while trajectories are better suited for scenarios involving tasks similar to those already completed.
By combining procedure knowledge from both methods of employing abstracted guidelines along with concrete execution trajectories, we attain the optimal performance.

After converting a set of completed trajectories into procedural memory, the next critical challenge is to retrieve the most accurate and relevant procedural knowledge when a new task arrives. We have designed several different key construction methods for memory storage to facilitate subsequent vector-based matching and retrieval:
\begin{itemize}
\item \textbf{Random Sample:} Does not utilize keys for vector retrieval; instead, randomly extracts a few memories from procedural memory.
\item \textbf{Query:} Employ query description as the key for storage, leveraging the semantic similarity of queries for retrieval.
\item \textbf{AveFact:} We apply a large model to extract keywords from the task's query, then computes the average similarity across matched keywords for retrieval.
\end{itemize}

During the retrieval process, we evaluate the similarity by calculating the cosine similarity between their corresponding vectors as shown in Table \ref{table: main}. 
Our experiments show that these different retrieval strategies produce varying results. Specifically, compared to random sampling, employing the query based and AveFact methods for precise retrieval significantly improves performance. 
The query-based approach benefits from capturing semantic contexts, enabling more accurate matches. The AveFact method, by extracting key features and averaging their similarities, effectively focuses on core task elements, leading to better retrieval efficacy. Overall, our findings suggest that incorporating semantic understanding and key feature extraction in retrieval strategies substantially enhances memory access accuracy and the effectiveness of downstream task performance.

\begin{table}[h]

\centering
\scalebox{0.7}{
\begin{tabular}{@{}c|l|ccc}
\toprule
\multirow{1}*{\textbf{Model}}&\multirow{1}*{\textbf{Policy}}
&\multicolumn{1}{c}{\#\textit{CS} ↑} &\multicolumn{1}{c}{\#\textit{HC} ↑} &\multicolumn{1}{c}{\multirow{1}*{{Steps ↓}}} \\
  % &  & \texttt{Micro} &  \texttt{Macro} &  \texttt{Micro} &  \texttt{Macro} & & \\
\midrule
\multirow{4}*{\textbf{GPT-4o}}&
No Memory & 71.93  & \textbf{12.88} & 17.84 \\
&Random Sample & \underline{74.59}   & 6.72  & \underline{15.12} \\
&Key=Query & 73.38  & \underline{8.95} &  15.44 \\
% &Key=FactString & 74.49 & 8.47  & 15.26 \\
&Key=AveFact & \textbf{76.02} & 8.25 & \textbf{14.64} \\
\midrule
\multirow{4}*{\textbf{Claude-3.5-sonnet}} &
No Memory & 63.49 & \textbf{33.06} & 18.84 \\
&Random Sample & 63.99 & \underline{29.91}  & 17.93 \\
&Key=Query & \underline{64.93}  & 28.56  & \textbf{17.60} \\
% &Key=FactString & 64.76  & 29.04 & 17.90 \\
&Key=AveFact & \textbf{65.76} & 29.61 & \underline{17.72} \\
\midrule
\multirow{4}*{\textbf{Qwen2.5-72b}} &
No Memory & 56.57  & 7.34 & 18.32 \\
&Random Sample & 59.76  & 8.43 & \underline{18.31} \\
&Key=Query & \underline{61.71} &  \underline{11.97} & 18.54 \\
% &Key=FactString & \underline{62.54} & \textbf{14.02}  & 18.59 \\
&Key=AveFact & \textbf{63.41} & \textbf{12.66} & \textbf{18.12} \\
\bottomrule
\end{tabular}
}
\caption{Results on \textbf{Retrieve Policy} on TravelPlanner.}
\label{table: main}
\end{table}

\begin{figure*}[!ht]
    \centering
    \scalebox{1}{
    \includegraphics[width=0.9\linewidth]{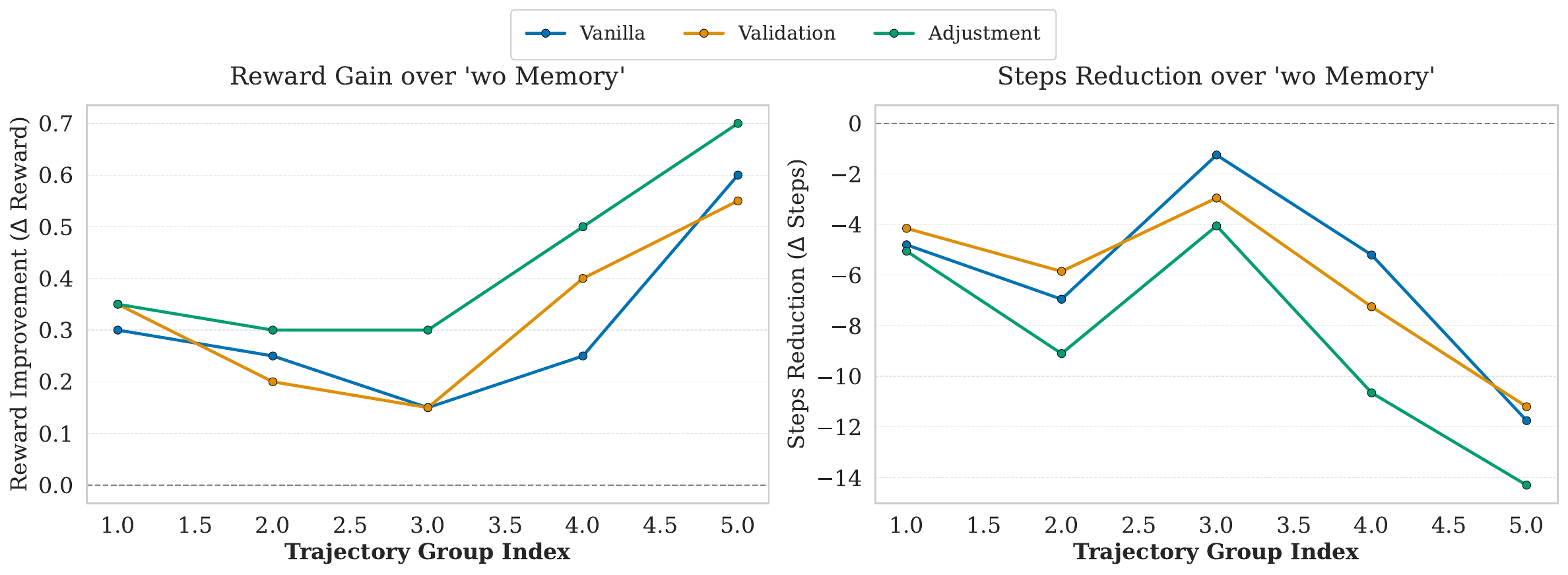} 
    }
    \caption{Reward gain and steps reduction vs. trajectory group index with \textbf{procedural memory}.}
    \label{fig:update}
    % \vspace{-1mm}
\end{figure*} 
\subsection{Memory Update}
% 很多现有的工作专注于如何建立一个可以复用的流程性知识, 使得模型不是独立的完成每个测试集上的任务, 而是learn from 之前的experience, 然而尽管这十分重要, 许多工作对于如何维护一个sequentially 运行的memory的动态更新的做法都十分初步, 要么使直接将新获得的memory 来增加meory的量(merge), 为此我们这里采用了多种memory online 更新的方式, 来探索那种动态memory更新策略在我们的任务上能够去得最好的效果. 并且除了端到端的指标, 我们也关注随着执行任务的数量的动态增加, procedural memory带来的accuracy 和effiency的提升
While many prior efforts have focused on developing reusable procedural knowledge, enabling models to learn from prior experiences rather than solving each test task in isolation, most existing memory update methods remain quite rudimentary. Typically, they simply append newly acquired memories to the existing store—a so-called ``merge'' strategy. In this work, we explore several online memory-update mechanisms to identify which dynamic strategy delivers the best performance on our tasks.
Beyond end-to-end evaluation metrics, we also analyze how both accuracy and efficiency evolve as the number of executed tasks increases, explicitly measuring the benefits conferred by our procedural memory.

To facilitate systematic comparison, we design several memory-update scenarios. 
The agent’s episodic memory is refreshed after every t test-set tasks, following the specific update strategies:
%The specific update strategies are as follows:
\begin{itemize}
\item \textbf{Vanilla Memory Update}:  
   After every $t$ tasks, all trajectories from these tasks are consolidated into procedural memories and directly appended to the memory bank.

\item \textbf{Validation}:  
    After every $t$ tasks, the system performs a selective consolidation step: only the trajectories that terminated in successful task completion are extracted, abstracted into compact, symbolic procedural memories. Trajectories that ended in failure, along with any redundant or noisy data, are discarded.
\item \textbf{Adjustment}:  
   When a retrieved procedural memory results in a failed execution, the erroneous trajectory is combined with the original memory and then revised in place, yielding an updated procedural memory.
\end{itemize}

% As illustrated in the Figure~\ref{fig:update}, we partitioned the tasks on the testbed into several groups, each containing numerous individual tasks. After completing one group, we applied the strategies above to build, store, and update the memory. The results show that, as more groups are finished and the memory is iteratively updated, every strategy yields higher scores on subsequent tasks while simultaneously reducing the number of steps required.
% Moreover, the strategies differ markedly in performance. 
% The reflexion-based update delivers the best outcome: on the final group of tasks it outperforms the second-best strategy by +0.7 points and saves 14 steps. This confirms that dynamically updating the memory, especially when coupled with an error-correction mechanism, continuously refines a high-quality procedural memory repository.
As depicted in Figure~\ref{fig:update}, we systematically divided the tasks within our testbed into several distinct groups, with each group comprising a diverse set of individual tasks. Upon the completion of tasks within each group, we employed the previously described strategies to construct, store, and update the procedural memory. The experimental results reveal a clear trend: as we sequentially progress through more groups and iteratively refresh the memory, all strategies contribute to improved performance on subsequent tasks. 

A closer comparison of different strategies exposes significant disparities in their effectiveness. Notably, the reflexion-based update mechanism stands out as the most effective approach. By the time the final group of tasks is reached, this method delivers a substantial advantage: it surpasses the second-best strategy by an impressive margin of +0.7 points and achieves a reduction of 14 steps. These improvements underscore the value of continually updating the memory, particularly when the update is guided by an error-correction mechanism embedded in the reflexion process.
\subsection{Comparison}
\begin{figure}[t] % !t, !htbp
    \centering
    \scalebox{1}{\includegraphics[width=1\linewidth]{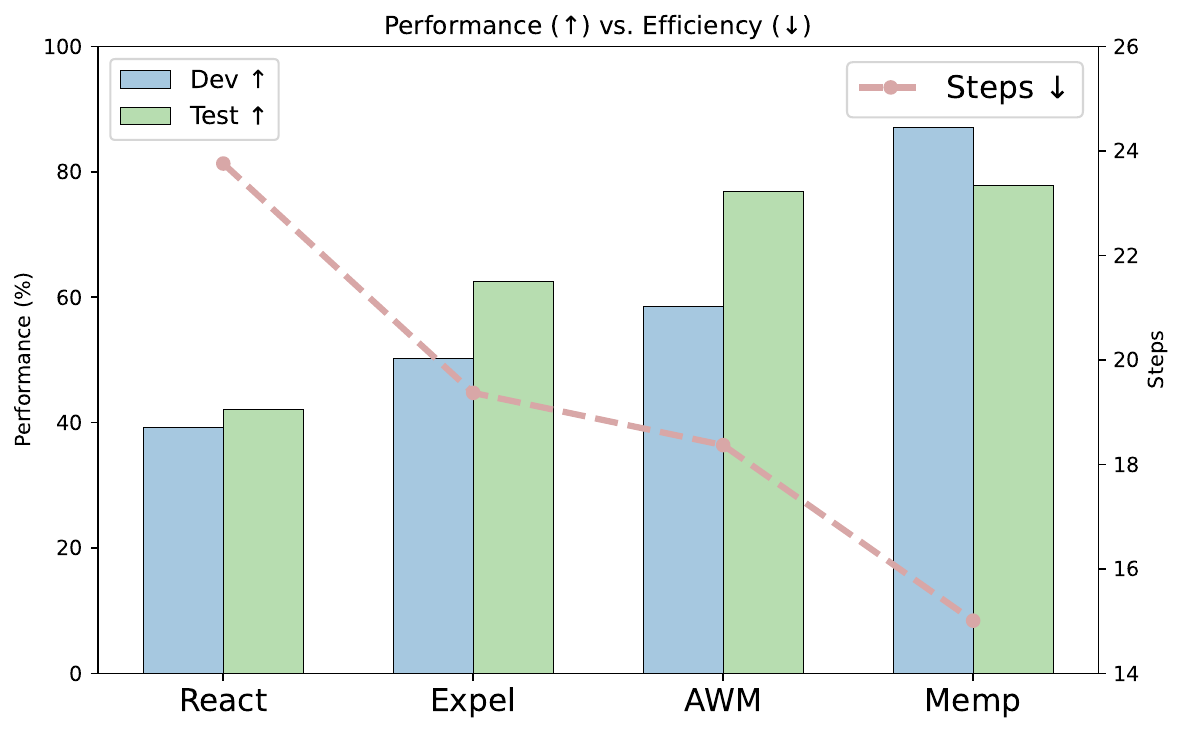} 
    }
    \caption{Performance of several baselines using GPT-4o on ALFWorld}
    \label{fig:memory compare}
    % \vspace{-1mm}
\end{figure} 
Figure~\ref{fig:memory compare} compares the performance and efficiency of different baselines using GPT-4o on ALFWorld. React performs the worst, with low success rates and the highest number of steps, indicating inefficient interaction. Expel and AWM achieve progressively better performance, with AWM outperforming Expel on both splits. $Mem^p$ delivers the best results, attaining the highest dev and test success rates while requiring the fewest steps. This shows that memory-augmented mechanisms significantly improve both task effectiveness and execution efficiency in long-horizon embodied agent tasks.
\section{Analysis}

\paragraph{Procedural Memory Boosts Accuracy and Cuts Trials.}
Figure~\ref{fig:case} presents a case study demonstrating how Procedural Memory enhances both accuracy and efficiency. 
In the absence of Procedural Memory, facing a complex task that has not been performed before, there are usually two situations.
In the first scenario, the model repeatedly attempts illegal or incorrect actions, causing the context to become increasingly complex and eventually exceeding the model's understanding capacity. 
In the second scenario, after multiple attempts, the model completes the task but at a cost significantly higher than the optimal path.

In contrast, once Procedural Memory is available for similar tasks, the model spends less time on trial and error. For example, in the egg heating problem, Procedural Memory can indicate the approximate location of the egg, saving the aimless search. 
During the heating process, it provides clear guidance, ensuring that the heating actions are performed consecutively and correctly, thereby allowing the task to be completed in fewer steps.

\begin{figure}[t] % !t, !htbp
    \centering
    \scalebox{1}{\includegraphics[width=1\linewidth]{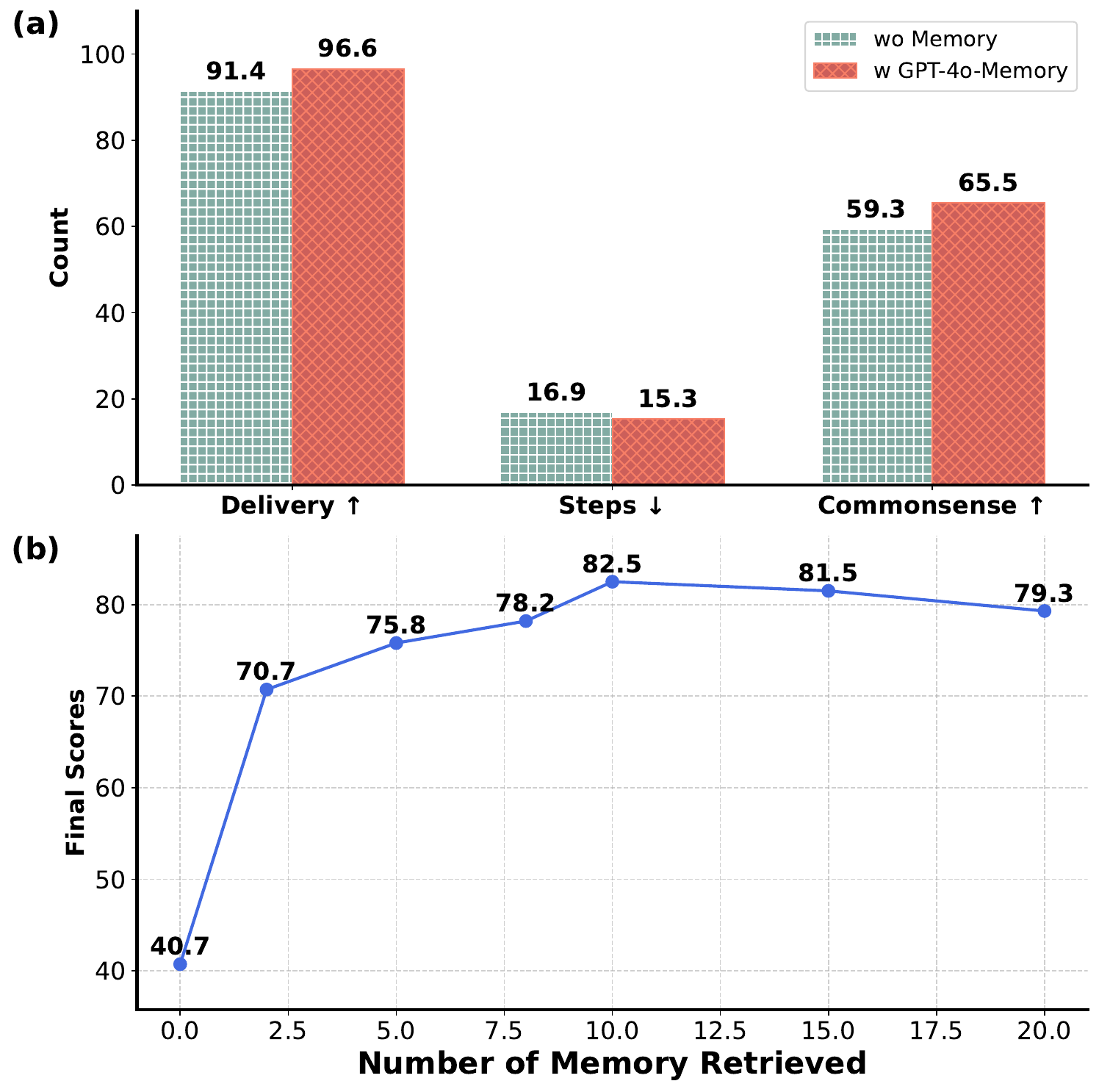} 
    }
    \caption{\textbf{(a)} Transfer result of GPT-4o's procedural memory to Qwen2.5-14B-Instruct and its performance on TravelPlanner dataset.\textbf{(b)} The relationship between the quantity of procedural memory retrieved for GPT-4o's performance on the ALFWorld dataset. }
    \label{fig:memory scaling}
    % \vspace{-1mm}
\end{figure} 

\paragraph{Procedural memory exhibits transferability from strong models to weaker ones.}
For a procedural memory constructed from a strong model in an offline memory library, we aim to verify whether this form of procedural memory can be effectively transferred to other models, or even weaker models. 
This exploration underscores the significance of memory transfer, as it could potentially enhance the adaptability and efficiency of various models by leveraging the knowledge and experience encapsulated within the strong model’s memory structure.
As shown in Figure~\ref{fig:memory scaling} (b), procedural memory generated by GPT-4o was employed by Qwen2.5-14B.
On the Travel Plan benchmark, the 14 billion parameter model raised its task completion rate by 5\% and cut the average number of steps by 1.6. Similar gains, both in success rate and trajectory length, appeared on ALFWorld. 
These outcomes confirm that procedural knowledge from a stronger model can be distilled into a reusable memory bank and transferred to a smaller model with minimal overhead, giving that smaller model a clear boost in task solving ability. 

\paragraph{Scaling Memory Retrieval Improves Agent Performance.}
While our main experiment has already demonstrated that procedural memory improves an agent’s task accuracy and reduces the number of steps required, vector-based storage and retrieval confer an advantage over human procedural memory: they can be scaled both in total capacity and in the number of memories retrieved. To investigate whether an agent’s performance continues to rise as the procedural-memory store and the number of retrieved memories increase, we designed a set of follow-up experiments.
As shown in Figure~\ref{fig:memory scaling} (b),  as the number of retrieved procedural memories increases, the agent's performance also improves steadily, exhibiting an upward trend followed by a plateau. However, retrieving too many memories can lead to a decline in the agent's performance. 
This is because excessive retrieval can affect the context length and also introduce less accurate procedural memories, which can interfere with the overall effectiveness.

% \subsection{Self Verify}
% \se

\section{Conclusion and Future Work}

We introduce $Mem^p$, a task-agnostic framework that elevates procedural memory to a core optimization target in LLM-based agents. By systematically studying strategies for memory construction, retrieval, and updating, $Mem^p$ enables agents to distill, reuse, and refine their own past experiences across diverse, long-horizon tasks and supports continual learning and robust generalization, marking a step toward self-improving, resilient agents.

In our experiments, $Mem^p$ has achieved promising results in both construction and retrieval. 
Moving forward, we plan to enhance this work in several ways.
Firstly, we will develop more diverse retrieval strategies. The current approach involves constructing different keys for vector-based retrieval. 
Secondly, in $Mem^p$, we currently rely on the standard reward signals provided by the benchmark. 
In such cases, using a LLM as a judge to assess task completion could be a viable solution. This would transform the agent's lifecycle into a continuous loop of executing tasks, self-assessing completion, building memories, and then progress.

%\newpage
\section*{Limitations}
$Mem^p$ is still constrained by two key limitations: 1. retrieval is restricted to vector-similarity search with manually crafted keys—other classic, potentially more precise methods like BM25 have not been incorporated; 2. the framework depends on explicit, benchmark supplied reward signals, so it cannot yet judge task success in real world settings where such rewards are sparse or absent.

\section*{Acknowledgement}
We would like to express sincere gratitude to the  reviewers for their thoughtful and constructive feedback. This work was supported by the National Natural Science Foundation of China (No. 62576307), Yongjiang Talent Introduction Programme (2021A-156-G), and Information Technology Center and State Key Lab of CAD\&CG, Zhejiang University.

%input{Sections/Reproduce.tex}

\bibliography{custom}

\clearpage
\appendix

\section*{Appendix}

\section{Use of AI Assistant}
We affirm that Large Language Models are employed solely as an assisted tool to refine wording
and sentence structure during our paper writing process. Their use in the experiments is strictly for
scientific research purposes, and all such usage has been explicitly documented in our Experimental
Settings and Reproducibility Statement. No other reliance on LLMs is involved in this work.

\section{Case Study}
\begin{figure*}[!h] % !t, !htbp
\vspace*{-15cm}
    \centering
    \scalebox{1}{
    \includegraphics[width=1\linewidth]{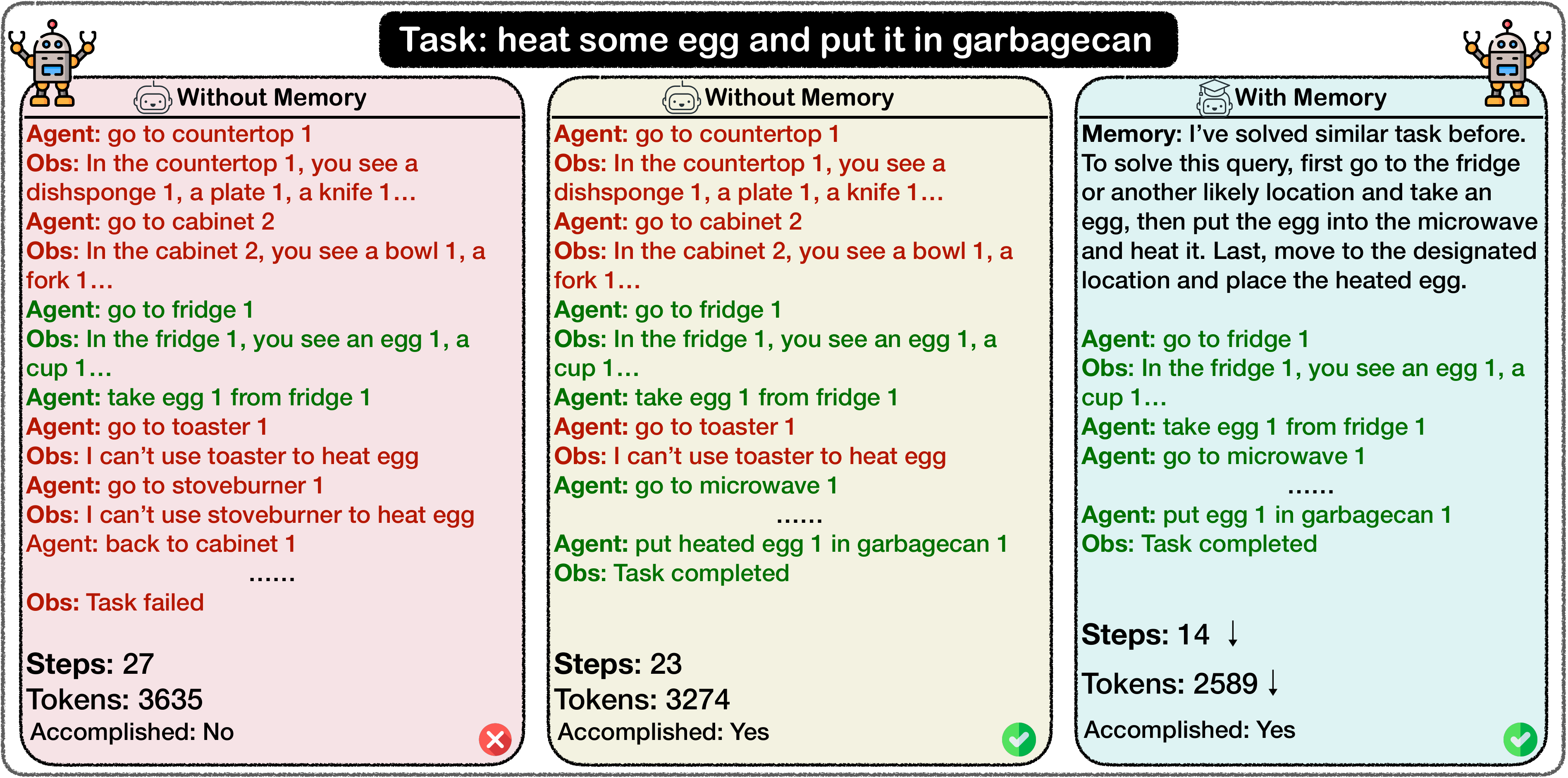} 
    }
    \caption{Compare trajectories with and without procedural memory, shortens the process by \textbf{9} steps and saves \textbf{685} tokens.}
    \label{fig:case}
    % \vspace{-1mm}
\end{figure*}

Detailed case studies are provided in Figure~\ref{fig:case}.

\section{Evaluation Details}
\label{app: evaluation}
For ALFWorld dataset, task completion is evaluated by the execution environment. 
After a task is completed or the maximum number of execution steps is reached, the environment provides a reward of 0 or 1 to indicate whether the task has been successfully completed.
For TravelPlanner, we conduct experiments on the test set in a two-stage mode.
After multiple rounds of interaction to obtain the travel trajectory and the final planner, GPT-4o converts the travel plan into a specified JSON format. 
The converted plan is then compared with the gold standard to obtain scores for both Common Sense and Hard Constraint.

\end{document}